\documentclass[11pt,a4paper]{article}
\usepackage[hyperref]{emnlp2018}
\usepackage{times}
\usepackage{latexsym}
\usepackage{graphicx}
\usepackage{multirow}
\usepackage{array}
\usepackage{multicol}
\usepackage{amsmath}

\usepackage{footnote}
\makesavenoteenv{tabular}
\makesavenoteenv{table}

\usepackage{url}

\aclfinalcopy 

\DeclareMathOperator*{\LSTMb}{LSTM_b}
\DeclareMathOperator*{\LSTMa}{LSTM_a}
\DeclareMathOperator*{\LSTMc}{LSTM_c}
\DeclareMathOperator*{\LSTMs}{Stack-LSTM}

\title{A Neural Transition-based Model for Nested Mention Recognition}

\author{
Bailin Wang \\
University of Massachusetts \\ Amherst \\
  {\tt bailinwang@cs.umass.edu}
  \And
Wei Lu  \\
Singapore University of Technology \\ and Design \\
  {\tt luwei@sutd.edu.sg} \\
  \AND
Yu Wang \and Hongxia Jin  \\
Samsung Research America \\
  {\tt \{yu.wang1, hongxia.jin\}@samsung.com}
  }

\date{}

\begin{document}
\maketitle
\begin{abstract}
It is common that entity mentions can contain other mentions recursively.
This paper introduces a scalable transition-based method to model the nested structure of mentions.
We first map a sentence with nested mentions to a designated forest where each mention corresponds to a constituent of the forest.
Our shift-reduce based system then learns to construct the forest structure in a bottom-up manner through an action sequence whose maximal length is guaranteed to be  three times of the sentence length.
Based on Stack-LSTM which is employed to efficiently and effectively represent the states of the system in a continuous space,
our system is further incorporated with a character-based component to capture letter-level patterns.
Our model achieves the state-of-the-art results on ACE datasets, showing its effectiveness in detecting nested mentions.\footnote{We make our implementation available at \url{https://github.com/berlino/nest-trans-em18}.}

\end{abstract}

\section{Introduction}

\begin{figure}[t]
\includegraphics[width=8cm]{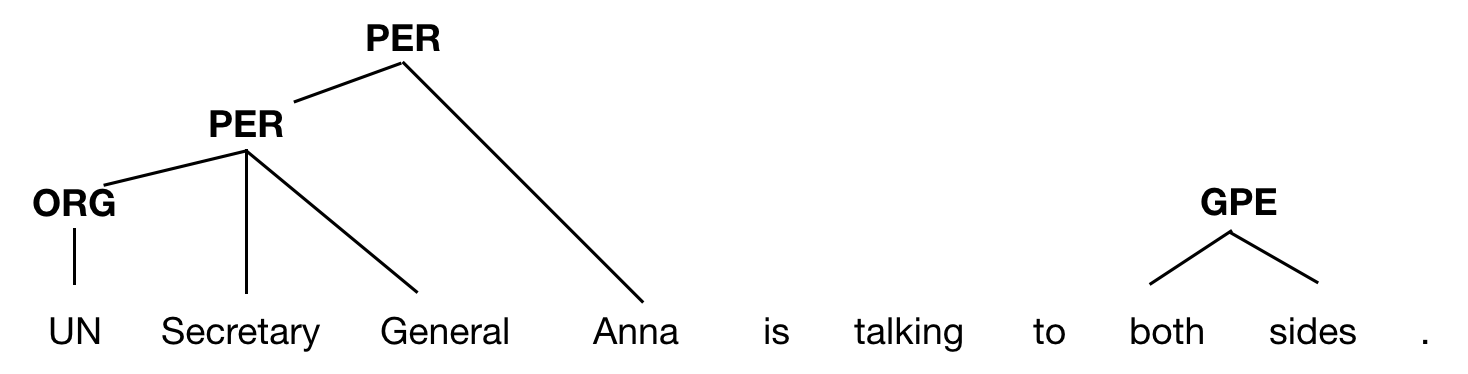}
\centering
\caption{An example sentence of nested mentions represented in the structure of forest. PER:Person, ORG:Organization, GPE:Geo-Political Entity. }
\label{fig:example}
\vspace{0mm}
\end{figure}

There has been an increasing interest in named entity recognition or more generally recognizing entity mentions\footnote{Mentions are defined as references to entities that could be named, nominal or pronominal \cite{florian2004statistical}.}
\cite{alex2007recognising,finkel2009nested,lu2015joint,muis2017labeling}
that the nested hierarchical structure of entity mentions should be taken into account to better facilitate downstream tasks like question answering \cite{abney2000answer}, relation extraction \cite{mintz2009distant,liu2017heterogeneous}, event extraction \cite{riedel2011fast,li-ji-huang:2013:ACL2013}, and coreference resolution \cite{soon2001machine,ng2002improving,chang2013constrained}.
Practically, the mentions with nested structures frequently exist in news \cite{doddington2004automatic} and  biomedical documents \cite{kim2003genia}.
For example in Figure \ref{fig:example},  ``UN Secretary General" of type Person also contains ``UN" of type Organization.

Traditional sequence labeling models such as conditional random fields (CRF) \cite{lafferty2001conditional} do not allow hierarchical structures between segments, making them incapable to handle such problems.
\citet{finkel2009nested} presented a chart-based parsing approach where each sentence with nested mentions is mapped to a rooted constituent tree.
The issue of using a chart-based parser is its cubic time complexity in the number of words in the sentence.

To achieve a scalable and effective solution for recognizing nested mentions, we design a transition-based system
which is inspired by the recent success of employing transition-based methods for constituent parsing \cite{zhang2009transition} and named entity recognition \cite{lou2017transition},
especially when they are paired with neural networks \cite{watanabe2015transition}.
Generally,  each sentence with nested mentions is mapped to a forest where each outermost mention forms a tree consisting of its inner mentions.
Then our transition-based system learns to construct this forest through a sequence of shift-reduce actions.
Figure \ref{fig:example} shows an example of such a forest.
In contrast, the tree structure by \citet{finkel2009nested} further uses a root node to connect all tree elements.
Our forest representation eliminates the root node so that the number of actions required to construct it can be reduced significantly.

Following \cite{P15-1033}, we employ Stack-LSTM  to represent the system's state, which consists of the states of input, stack and action history, in a continuous space incrementally.
The (partially) processed nested mentions in the stack are encoded with recursive neural networks \cite{socher2013recursive}
where composition functions are used to capture dependencies between nested mentions.
Based on the observation that letter-level patterns such as capitalization and prefix can be beneficial in detecting mentions, we incorporate a character-level LSTM to capture such morphological information.
Meanwhile, this character-level component can also help deal with the out-of-vocabulary problem of neural models.
We conduct experiments in three standard datasets.
Our system achieves the state-of-the-art performance on ACE datasets and comparable performance in GENIA dataset.


\section{Related Work}

Entity mention recognition with nested structures has been explored first with rule-based approaches \cite{zhang2004enhancing,zhou2004recognizing,zhou2006recognizing}
where the authors first detected the innermost mentions and then relied on rule-based post-processing methods to identify outer mentions.
\citet{mcdonald2005flexible} proposed a structured multi-label model to represent overlapping segments in a sentence.
but it came with a cubic time complexity in the number of words.
\citet{alex2007recognising} proposed several ways to combine multiple conditional random fields (CRF) \cite{lafferty2001conditional} for such tasks.
Their best results were obtained by cascading several CRF models in a specific order while each model is responsible for detecting mentions of a particular type.
However, such an approach cannot model nested mentions of the same type, which frequently appear.

\citet{lu2015joint} and \citet{muis2017labeling}  proposed new representations of mention hypergraph and mention separator to model {\em overlapping mentions}.
However, the nested structure is not guaranteed in such approaches since overlapping structures additionally include the {\em crossing structures}\footnote{For example, in a four-word sentence ABCD, the phrase ABC and BCD together form a {crossing structure}.}, which rarely exist in practice \cite{lu2015joint}.
{\color{black}Also, their representations did not model the dependencies between nested mentions explicitly, which may limit their performance.}
In contrast, the chart-based parsing method \cite{finkel2009nested} can capture {\color{black} the dependencies between nested mentions with composition rules which allow an outer entity to be influenced by its contained entities}.
However, their cubic time complexity makes them not scalable to large datasets.

As neural network based approaches are proven  effective in entity or mention recognition \cite{collobert2011natural,lample2016neural,huang2015bidirectional,chiu2016named,ma-hovy:2016:P16-1},
recent efforts focus on incorporating neural components for recognizing nested mentions.
\citet{N18-1131} dynamically stacked multiple LSTM-CRF layers \cite{lample2016neural}, detecting mentions in an inside-out manner
until no outer entities are extracted.
\citet{N18-1079} used recurrent neural networks to extract features for a hypergraph which encodes all nested mentions based on the BILOU tagging scheme.

\section{Model}

Specifically, given a sequence of words $\lbrace x_0, x_1, \dots, x_n\rbrace$, the goal of our system is to output a set of mentions where nested structures are allowed.
We use the forest structure to model the nested mentions scattered in a sentence, as shown in Figure \ref{fig:example}.
The mapping is straightforward: each outermost mention forms a tree
where the mention is the root and its contained mentions correspond to constituents of the tree.\footnote{Note that words that are not contained in any mention each forms a single-node tree.}

\subsection{Shift-Reduce System}

\begin{figure}[t!]
\begin{center}
\renewcommand{\arraystretch}{0.8}
\begin{tabular}{>{\small}c>{\small}c}
Initial State & $[\phi, 0, \phi]$\\
Final State & $[ S, n, A]$\\
\\
\textsc{Shift} & {\Large$\frac{[S, \ i, \ A]}{[S|w, \ i+1, \ A|\textsc{Shift}]}$} \\
\\
\textsc{Reduce-X} & {\Large$\frac{[S|t_1t_0, \ i, \ A]}{[S|X, \ i, \ A|\textsc{Reduce-X}]}$} \\
\\
\textsc{Unary-X} & {\Large$\frac{[S|t_0, \  i,  \ A]}{[S|X, \ i,  \ A|\textsc{Unary-X}]}$} \\
\\
\end{tabular}
\end{center}
\caption{\label{fig:ds1} Deduction rules. $[ S, i, A]$ denotes stack, buffer front index and action history respectively. }
\end{figure}

Our transition-based model is based on the shift-reduce parser for constituency parsing \cite{watanabe2015transition}, which adopts \cite{zhang2009transition,sagae2005classifier}.
Generally, our system employs a stack to store (partially) processed nested elements.
The system's state is defined as $[S, i, A]$ which denotes stack, buffer front index and action history respectively.
In each step. an action is applied to change the system's state.

Our system consists of three types of transition actions, which are also summarized in Figure \ref{fig:ds1}:
\begin{itemize}
  \item \textsc{Shift} pushes the next word from buffer to the stack.
  \item  \textsc{Reduce-X}  pops the top two items $t_0$ and $t_1$ from the tack and combines them as a new tree element \{$\textsc{X} \rightarrow t_0t_1$\} which is then pushed onto the stack.
  \item  \textsc{Unary-X} pops the top item $t_0$ from the stack and constructs a new tree element \{$\textsc{X} \rightarrow t_0$\} which is pushed back to the stack.
\end{itemize}


Since the shift-reduce system assumes unary and binary branching, we binarize the trees in each forest in a left-branching manner.
For example, if three consecutive words $A, B, C$ are annotated as Person, we convert it into a binary tree $\{Person \rightarrow \{ Person* \rightarrow A, B \}, C\}$
where $Person*$ is a temporary label for $Person$.
Hence, the $\textsc{X}$ in reduce- actions will also include such temporary labels.

\begin{figure}[t]
\includegraphics[width=\linewidth, height=9.2cm]{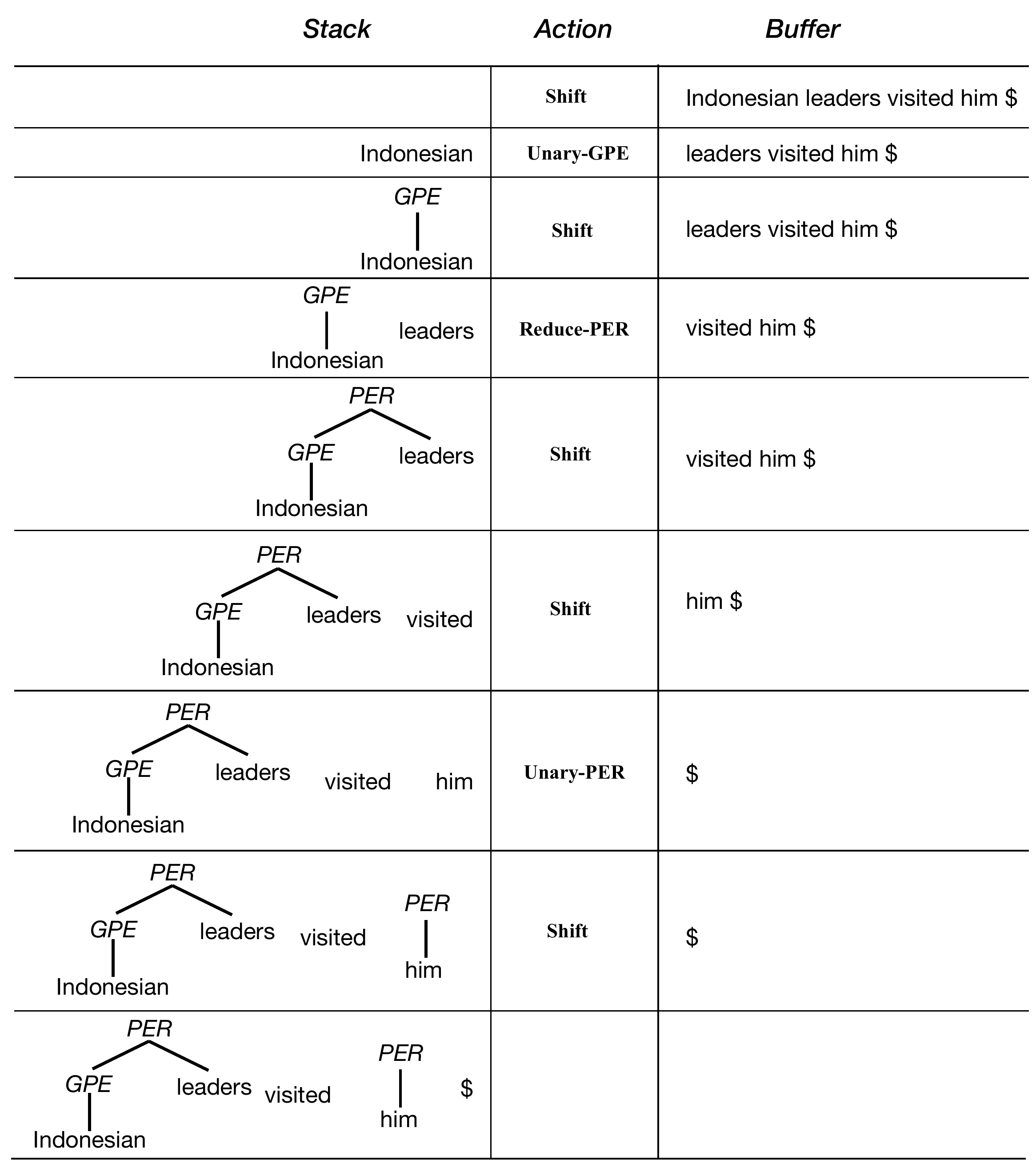}
\centering
\caption{An example sequence of transition actions for the sentence ``Indonesian leaders visited him''.
\$ is the special symbol indicating the termination of transitions. PER:Person, GPE:Geo-Political Entity.}
\label{fig:transitions}
\end{figure}

Note that since most words are not contained in any mention, they are only shifted to the stack and not involved in any reduce- or unary- actions.
An example sequence of transitions can be found in Figure \ref{fig:transitions}.
Our shift-reduce system is different from previous parsers in terms of the terminal state.
1) It does not require the terminal stack to be a rooted tree.
Instead, the final stack should be a forest consisting of multiple nested elements with tree structures.
2) To conveniently determine the ending of our transition process, we add an auxiliary symbol $\$$ to each sentence.
Once it is pushed to the stack, it implies that all deductions of actual words are finished.
Since we do not allow unary rules between labels like $\textsc{X1} \rightarrow \textsc{X2}$,
the length of maximal action sequence is $3n$.\footnote{In this case, each word is shifted ($n$) and involved in a unary action ($n$). Then all elements are reduced to a single node ($n-1$).
The last action is to shift the symbol $\$$.  }

\subsection{Action Constraints}

To make sure that each action sequence is valid,
we need to make some hard constraints on the action to take.
For example, reduce- action can only be conducted when there are at least two elements in the stack.
Please see the Appendix for the full list of restrictions.
Formally, we use $\mathcal{V}(S,i,A)$ to denote the valid actions given the parser state.
Let us denote the feature vector for the parser state at time step $k$ as $\mathbf{p}_k$.
The distribution of actions is computed as follows:
\begin{equation}
p(z_k \mid \mathbf{p}_k) = \frac{\exp \left( \mathbf{w}_{z_k}^{\top} \mathbf{p}_k + b_{z_k} \right)}{\sum_{z' \in \mathcal{V}(S,i,A)} \exp \left( \mathbf{w}_{z'}^{\top} \mathbf{p}_k + b_{z'} \right)}
\label{eq:probs}
\end{equation}

\noindent where $\mathbf{w}_z$ is a column weight vector for action $z$, and $b_z$ is a bias term.

\subsection{Neural Transition-based Model}

We use neural networks to learn the representation of the parser state, which is $\mathbf{p}_k$ in (\ref{eq:probs}).

\subsubsection*{Representation of Words}
\label{sec:wr}

Words are represented by concatenating three vectors:
\begin{equation}
\mathbf e_{x_i} = [ \mathbf e_{w_i}, \mathbf e_{p_i}, \mathbf c_{w_i}]
\label{eq:nmr_x}
\end{equation}

\noindent where $ \mathbf e_{w_i} $ and $\mathbf e_{p_i}$ denote the embeddings for $i$-th word and its POS tag respectively.
$\mathbf c_{w_i}$ denotes the representation learned by a character-level model using a bidirectional LSTM.
Specifically, for character sequence $s_0, s_1,\dots, s_n$ in the $i$-{th} word, we use the last hidden states of forward and backward LSTM as the character-based representation of this word, as shown below:
\begin{equation}
\mathbf c_{w_i} = [  \overrightarrow \LSTMc (s_0, \dots, s_n), \overleftarrow \LSTMc (s_0, \dots, s_n) ]
\label{eq:char_x}
\end{equation}

\subsubsection*{Representation of Parser States}

Generally, the buffer and action history are encoded using two vanilla LSTMs \cite{graves2005framewise}.
For the stack that involves popping out top elements, we use the Stack-LSTM \cite{P15-1033} to efficiently encode it.

Formally, if the unprocessed word sequence in the buffer is $x_i, x_{i+1},\dots, x_{n}$ and action history sequence is $a_0, a_{1},\dots,a_{k-1}$, then we can
compute buffer representation $\mathbf{b}_k$ and action history representation $\mathbf{a}_k$ at time step $k$ as follows:
\begin{align}
\mathbf{b}_k = \overleftarrow \LSTMb [\mathbf{e}_{x_i}, \dots, \mathbf{e}_{x_{n}} ] \ \ \ \\ \label{eq:buffer}
\mathbf{a}_k = \overrightarrow \LSTMa[\mathbf{e}_{a_0}, \dots, \mathbf{e}_{a_{k-1}} ]
\end{align}
where each action is also mapped to a distributed representation  $\mathbf{e}_{a}$.\footnote{Note that $\LSTMb$ runs in a right-to-left order such that the output can represent the contextual information of $x_i$.}
For the state of the stack, we also use an LSTM to encode a sequence of tree elements.
However, the top elements of the stack are updated frequently.
Stack-LSTM provides an efficient implementation that incorporates a stack-pointer.\footnote{Please refer to \citet{P15-1033} for details.}
Formally, the state of the stack $\mathbf b_k$ at time step $k$ is computed as:
\begin{equation}
\mathbf{s}_k = \LSTMs [\mathbf{h}_{t_m}, \dots, \mathbf{h}_{t_0} ]
\end{equation}
where $\mathbf{h}_{t_i}$ denotes the representation of the $i$-{th} tree element from the top,
which can be computed recursively similar to Recursive Neural Network \cite{socher2013recursive} as follows:
\begin{align}
 \mathbf{h}_{parent} =  \mathbf{W}_{u,l}^{\top} \mathbf{h}_{child} + \mathbf{b}_{u,l}  \quad \quad  \quad \quad \ \\
 \mathbf{h}_{parent} =  \mathbf{W}_{b,l}^{\top} [\mathbf{h}_{lchild}, \mathbf{h}_{rchild} ] + \mathbf{b}_{u,l} \ \
\end{align}
where $\mathbf{W}_{u,l}$ and $\mathbf{W}_{b,l}$ denote the weight matrices for unary($u$) and binary($b$) composition with parent node being label($l$).
Note that the composition function is distinct for each label $l$.
Recall that the leaf nodes of each tree element are raw words.
Instead of representing them with their original embeddings introduced in Section \ref{sec:wr},
we found that concatenating the buffer state in (\ref{eq:buffer})  are beneficial during our initial experiments.
Formally, when a word $x_i$ is shifted to the stack at time step $k$, its representation is computed as:
\begin{equation}
\mathbf{h}_{leaf} =  \mathbf{W}_{leaf}^{\top} [ \mathbf e_{x_i}, \mathbf b_k ] + \mathbf{b}_{leaf}
\end{equation}

Finally, the state of the system $\mathbf p_k$ is the concatenation of the states of buffer, stack and action history:
\begin{equation}
 \mathbf{p}_{k} = [\mathbf b_k, \mathbf s_k, \mathbf a_k]
\end{equation}

\subsubsection*{Training}

We employ the greedy strategy to maximize the log-likelihood of the local action classifier in (\ref{eq:probs}).
Specifically, let $z_{ik}$ denote the $k$-{th} action for the $i$-{th} sentence, the loss function with $\ell_2$ norm is:
\begin{equation}
 \mathcal L(\theta) =  - \sum_i \sum_k \log p(z_{ik}) + \frac{\lambda}{2} \Vert \theta \Vert ^2
\end{equation}
where $\lambda$ is the $\ell_2$ coefficient.

\begin{table}[t!]
\centering
\scalebox{0.72}
{
\begin{tabular}{l|c|c|c|c}
Models & ACE04 & ACE05 & GENIA & $w/s$ \\
  \hline
 \citet{finkel2009nested} & - & - & 70.3 & 38$^\dagger$ \\
 \citet{lu2015joint}   & 62.8  & 62.5 & 70.3  & 454  \\
\citet{muis2017labeling}   & 64.5  & 63.1 & 70.8  & 263   \\
\citet{N18-1079} & 72.7 & 70.5 & 73.6 & -\\
\citet{N18-1131}
\footnote{Note that in ACE2005, \citet{N18-1131} did their experiments with a different split from  \citet{lu2015joint} and \citet{muis2017labeling} which we follow as our split. }
 &  - & 72.2 & \bf{74.7} & -\\
\hline
Ours & \bf{73.3} & \bf{73.0} & 73.9 &  1445    \\
- char-level LSTM & 72.3 & 71.9 & 72.1 & 1546     \\
- pre-trained embeddings & 71.3 & 71.5 & 72.0 & 1452 \\
- dropout layer & 71.7 & 72.0 & 72.7 & 1440

\end{tabular}
}
\caption{Main results in terms of $F_1$ score (\%).
$w/s$: \# of words decoded per second,
number with $\dagger$ is retrieved from the original paper.
}
\label{tab:result}
\end{table}

\section{Experiments}

We mainly evaluate our models on the standard ACE-04, ACE-05 \cite{doddington2004automatic}, and GENIA \cite{kim2003genia} datasets  with the same splits used by previous research efforts \cite{lu2015joint,muis2017labeling}.
In ACE datasets, more than 40\% of the mentions form nested structures with some other mention.
In GENIA, this number is 18\%.
Please see \citet{lu2015joint} for the full statistics.

\subsection{Setup}

Pre-trained embeddings GloVe \cite{pennington2014glove} of dimension 100 are used to initialize the word vectors for all three datasets.\footnote{We also additionally tried using embeddings trained on PubMed for GENIA but the performance was comparable.}
The embeddings of POS tags are initialized randomly with dimension 32.
The model is trained using Adam \cite{kingma2014adam} and a gradient clipping of 3.0.
Early stopping is used based on the performance of development sets.
Dropout \cite{srivastava2014dropout} is used after the input layer.
The $\ell_2$ coefficient $\lambda$ is also tuned  during development process.

\subsection{Results}

The main results are reported in Table \ref{tab:result}.
Our neural transition-based model achieves the best results in ACE datasets and comparable results in GENIA dataset in terms of $F_1$ measure.
We hypothesize that the performance gain of our model compared with other methods is largely due to improved performance on the portions of nested mentions in our datasets.
To verify this, we design an experiment to evaluate how well a system can recognize nested mentions.

\subsubsection*{Handling Nested Mentions}

The idea is that we split the test data into two portions:
sentences with and without nested mentions.
The results of
GENIA are listed in Table \ref{tab:nested}.
We can observe that the margin of improvement is more significant in the portion of nested mentions, revealing our model's effectiveness in handling nested mentions.
This observation helps explain why our model achieves greater improvement in ACE than in GENIA in Table \ref{tab:result}  since the former has much more nested structures than the latter.
Moreover, \citet{N18-1131} performs better when it comes to non-nested mentions possibly due to the CRF they used, which globally normalizes each stacked layer.

\subsubsection*{Decoding Speed}

Note that \citet{lu2015joint} and \citet{muis2017labeling} also feature linear-time complexity, but with a greater constant factor.
To compare the decoding speed, we re-implemented their model with the same platform (PyTorch) and run them on the same machine (CPU: Intel i5 2.7GHz).
Our model turns out to be around 3-5 times faster than theirs, showing its scalability.

\subsubsection*{Ablation Study}

To evaluate the contribution of neural components including pre-trained embeddings, the character-level LSTM and dropout layers, we test the performances of ablated models.
The results are listed in Table \ref{tab:result}.
From the performance gap, we can conclude that these components contribute significantly to the effectiveness of our model in all three datasets.


\begin{table}[t!]
\centering
\scalebox{0.75}
{
%
\begin{tabular}{l|ccc|ccc}
 &  \multicolumn{6}{c}{GENIA} \\
    &  \multicolumn{3}{c|}{{Nested}} & \multicolumn{3}{c}{{Non-Nested}}  \\
              &$P$ & $R$ & $F_1$ & $P$ & $R$ & $F_1$   \\
 \hline
\citet{lu2015joint}             &  76.3 & 60.8 & 67.7 & 73.1 & 70.7 & 71.9   \\
\citet{muis2017labeling}          & 76.5 & 60.3 & 67.4 & 74.8 & 71.3 & 73.0  \\
\citet{N18-1131} & 79.4 & 63.6 & 70.6 & 78.5 & 77.5 & 78.0\\
\hline
Ours  &  80.3 & 64.6 & 71.6 & 76.8 & 73.9 & 75.3   \\
\end{tabular}
}
\caption{Results (\%) on different types of sentences on the GENIA dataset.}
\label{tab:nested}
\end{table}

\section{Conclusion and Future Work}

In this paper, we present a transition-based model for nested mention recognition using a forest representation.
Coupled with Stack-LSTM for representing the system's state, our neural model can capture dependencies between nested mentions efficiently.
Moreover, the character-based component helps capture letter-level patterns in words.
The system achieves the state-of-the-art performance in ACE datasets.

One potential drawback of the system is the greedy training and decoding.
We believe that alternatives like beam search and training with exploration \cite{goldberg2012dynamic} could further boost the performance.
Another direction that we plan to work on is to apply this model to recognizing overlapping and entities that involve discontinuous spans \cite{muis2016learning} which frequently exist in the biomedical domain.

\section*{Acknowledgements}

We would like to thank the anonymous reviewers for their
valuable comments.
We also thank Meizhi Ju for providing raw predictions and helpful discussions.
This work was done after the first author visited Singapore University of Technology and Design.
This work is supported by Singapore Ministry of Education Academic Research Fund (AcRF) Tier 2 Project MOE2017-T2-1-156.

\bibliography{transition}

\begin{thebibliography}{39}
\expandafter\ifx\csname natexlab\endcsname\relax\def\natexlab#1{#1}\fi

\bibitem[{Abney et~al.(2000)Abney, Collins, and Singhal}]{abney2000answer}
Steven Abney, Michael Collins, and Amit Singhal. 2000.
\newblock Answer extraction.
\newblock In \emph{Proc. of the sixth conference on applied natural language
  processing}.

\bibitem[{Alex et~al.(2007)Alex, Haddow, and Grover}]{alex2007recognising}
Beatrice Alex, Barry Haddow, and Claire Grover. 2007.
\newblock Recognising nested named entities in biomedical text.
\newblock In \emph{Proc. of BioNLP}.

\bibitem[{Chang et~al.(2013)Chang, Samdani, and Roth}]{chang2013constrained}
Kai-Wei Chang, Rajhans Samdani, and Dan Roth. 2013.
\newblock A constrained latent variable model for coreference resolution.
\newblock In \emph{Proc. of EMNLP}.

\bibitem[{Chiu and Nichols(2016)}]{chiu2016named}
Jason~PC Chiu and Eric Nichols. 2016.
\newblock Named entity recognition with bidirectional lstm-cnns.
\newblock \emph{TACL}.

\bibitem[{Collobert et~al.(2011)Collobert, Weston, Bottou, Karlen, Kavukcuoglu,
  and Kuksa}]{collobert2011natural}
Ronan Collobert, Jason Weston, L{\'e}on Bottou, Michael Karlen, Koray
  Kavukcuoglu, and Pavel Kuksa. 2011.
\newblock Natural language processing (almost) from scratch.
\newblock \emph{JMLR}.

\bibitem[{Doddington et~al.(2004)Doddington, Mitchell, Przybocki, Ramshaw,
  Strassel, and Weischedel}]{doddington2004automatic}
George~R Doddington, Alexis Mitchell, Mark~A Przybocki, Lance~A Ramshaw,
  Stephanie Strassel, and Ralph~M Weischedel. 2004.
\newblock The automatic content extraction (ace) program-tasks, data, and
  evaluation.
\newblock In \emph{Proc. of LREC}.

\bibitem[{Dyer et~al.(2015)Dyer, Ballesteros, Ling, Matthews, and
  Smith}]{P15-1033}
Chris Dyer, Miguel Ballesteros, Wang Ling, Austin Matthews, and Noah~A. Smith.
  2015.
\newblock Transition-based dependency parsing with stack long short-term
  memory.
\newblock In \emph{Proc. of ACL}.

\bibitem[{Finkel and Manning(2009)}]{finkel2009nested}
Jenny~Rose Finkel and Christopher~D Manning. 2009.
\newblock Nested named entity recognition.
\newblock In \emph{Proc. of EMNLP}.

\bibitem[{Florian et~al.(2004)Florian, Hassan, Ittycheriah, Jing, Kambhatla,
  Luo, Nicolov, and Roukos}]{florian2004statistical}
R.~Florian, H.~Hassan, A.~Ittycheriah, H.~Jing, N.~Kambhatla, X.~Luo,
  N.~Nicolov, and S.~Roukos. 2004.
\newblock A statistical model for multilingual entity detection and tracking.
\newblock In \emph{Proc. of HLT-NAACL}.

\bibitem[{Goldberg and Nivre(2012)}]{goldberg2012dynamic}
Yoav Goldberg and Joakim Nivre. 2012.
\newblock A dynamic oracle for arc-eager dependency parsing.
\newblock \emph{Proceedings of COLING 2012}, pages 959--976.

\bibitem[{Graves and Schmidhuber(2005)}]{graves2005framewise}
Alex Graves and J{\"u}rgen Schmidhuber. 2005.
\newblock Framewise phoneme classification with bidirectional lstm and other
  neural network architectures.
\newblock \emph{Neural Networks}.

\bibitem[{Huang et~al.(2015)Huang, Xu, and Yu}]{huang2015bidirectional}
Zhiheng Huang, Wei Xu, and Kai Yu. 2015.
\newblock Bidirectional lstm-crf models for sequence tagging.
\newblock \emph{arXiv preprint arXiv:1508.01991}.

\bibitem[{Ju et~al.(2018)Ju, Miwa, and Ananiadou}]{N18-1131}
Meizhi Ju, Makoto Miwa, and Sophia Ananiadou. 2018.
\newblock A neural layered model for nested named entity recognition.
\newblock In \emph{Proc. of NAACL-HLT}.

\bibitem[{Katiyar and Cardie(2018)}]{N18-1079}
Arzoo Katiyar and Claire Cardie. 2018.
\newblock Nested named entity recognition revisited.
\newblock In \emph{Proc. of NAACL-HLT}.

\bibitem[{Kim et~al.(2003)Kim, Ohta, Tateisi, and Tsujii}]{kim2003genia}
J-D Kim, Tomoko Ohta, Yuka Tateisi, and Jun'ichi Tsujii. 2003.
\newblock Genia corpus---a semantically annotated corpus for bio-textmining.
\newblock \emph{Bioinformatics}.

\bibitem[{Kingma and Ba(2014)}]{kingma2014adam}
Diederik Kingma and Jimmy Ba. 2014.
\newblock Adam: A method for stochastic optimization.
\newblock In \emph{Proc. of ICLR}.

\bibitem[{Lafferty et~al.(2001)Lafferty, McCallum, and
  Pereira}]{lafferty2001conditional}
John Lafferty, Andrew McCallum, and Fernando~CN Pereira. 2001.
\newblock Conditional random fields: Probabilistic models for segmenting and
  labeling sequence data.
\newblock In \emph{Proc. of ICML}.

\bibitem[{Lample et~al.(2016)Lample, Ballesteros, Subramanian, Kawakami, and
  Dyer}]{lample2016neural}
Guillaume Lample, Miguel Ballesteros, Sandeep Subramanian, Kazuya Kawakami, and
  Chris Dyer. 2016.
\newblock Neural architectures for named entity recognition.
\newblock In \emph{Proc. of NAACL-HLT}.

\bibitem[{Li et~al.(2013)Li, Ji, and Huang}]{li-ji-huang:2013:ACL2013}
Qi~Li, Heng Ji, and Liang Huang. 2013.
\newblock Joint event extraction via structured prediction with global
  features.
\newblock In \emph{Proc. of ACL}.

\bibitem[{Liu et~al.(2017)Liu, Ren, Zhu, Zhi, Gui, Ji, and
  Han}]{liu2017heterogeneous}
Liyuan Liu, Xiang Ren, Qi~Zhu, Shi Zhi, Huan Gui, Heng Ji, and Jiawei Han.
  2017.
\newblock Heterogeneous supervision for relation extraction: A representation
  learning approach.
\newblock In \emph{Proc. of EMNLP}.

\bibitem[{Lou et~al.(2017)Lou, Zhang, Qian, Li, Xiong, and
  Ji}]{lou2017transition}
Yinxia Lou, Yue Zhang, Tao Qian, Fei Li, Shufeng Xiong, and Donghong Ji. 2017.
\newblock A transition-based joint model for disease named entity recognition
  and normalization.
\newblock \emph{Bioinformatics}.

\bibitem[{Lu and Roth(2015)}]{lu2015joint}
Wei Lu and Dan Roth. 2015.
\newblock Joint mention extraction and classification with mention hypergraphs.
\newblock In \emph{Proc. of EMNLP}.

\bibitem[{Ma and Hovy(2016)}]{ma-hovy:2016:P16-1}
Xuezhe Ma and Eduard Hovy. 2016.
\newblock End-to-end sequence labeling via bi-directional lstm-cnns-crf.
\newblock In \emph{Proc. of ACL}.

\bibitem[{McDonald et~al.(2005)McDonald, Crammer, and
  Pereira}]{mcdonald2005flexible}
Ryan McDonald, Koby Crammer, and Fernando Pereira. 2005.
\newblock Flexible text segmentation with structured multilabel classification.
\newblock In \emph{Proc. of HLT-EMNLP}.

\bibitem[{Mintz et~al.(2009)Mintz, Bills, Snow, and
  Jurafsky}]{mintz2009distant}
Mike Mintz, Steven Bills, Rion Snow, and Dan Jurafsky. 2009.
\newblock Distant supervision for relation extraction without labeled data.
\newblock In \emph{Proc. of ACL-IJCNLP}.

\bibitem[{Muis and Lu(2016)}]{muis2016learning}
Aldrian~Obaja Muis and Wei Lu. 2016.
\newblock Learning to recognize discontiguous entities.
\newblock In \emph{Proceedings of the 2016 Conference on Empirical Methods in
  Natural Language Processing}, pages 75--84.

\bibitem[{Muis and Lu(2017)}]{muis2017labeling}
Aldrian~Obaja Muis and Wei Lu. 2017.
\newblock Labeling gaps between words: Recognizing overlapping mentions with
  mention separators.
\newblock In \emph{Proc. of EMNLP}.

\bibitem[{Ng and Cardie(2002)}]{ng2002improving}
Vincent Ng and Claire Cardie. 2002.
\newblock Improving machine learning approaches to coreference resolution.
\newblock In \emph{Proc. of ACL}.

\bibitem[{Pennington et~al.(2014)Pennington, Socher, and
  Manning}]{pennington2014glove}
Jeffrey Pennington, Richard Socher, and Christopher Manning. 2014.
\newblock Glove: Global vectors for word representation.
\newblock In \emph{Proc. of EMNLP}.

\bibitem[{Riedel and McCallum(2011)}]{riedel2011fast}
S.~Riedel and A.~McCallum. 2011.
\newblock {Fast and robust joint models for biomedical event extraction}.
\newblock In \emph{{Proc. of EMNLP}}.

\bibitem[{Sagae and Lavie(2005)}]{sagae2005classifier}
Kenji Sagae and Alon Lavie. 2005.
\newblock A classifier-based parser with linear run-time complexity.
\newblock In \emph{IWPT}.

\bibitem[{Socher et~al.(2013)Socher, Perelygin, Wu, Chuang, Manning, Ng, and
  Potts}]{socher2013recursive}
Richard Socher, Alex Perelygin, Jean Wu, Jason Chuang, Christopher~D Manning,
  Andrew Ng, and Christopher Potts. 2013.
\newblock Recursive deep models for semantic compositionality over a sentiment
  treebank.
\newblock In \emph{Proc. of EMNLP}.

\bibitem[{Soon et~al.(2001)Soon, Ng, and Lim}]{soon2001machine}
Wee~Meng Soon, Hwee~Tou Ng, and Daniel Chung~Yong Lim. 2001.
\newblock A machine learning approach to coreference resolution of noun
  phrases.
\newblock \emph{Computational linguistics}, 27(4):521--544.

\bibitem[{Srivastava et~al.(2014)Srivastava, Hinton, Krizhevsky, Sutskever, and
  Salakhutdinov}]{srivastava2014dropout}
Nitish Srivastava, Geoffrey Hinton, Alex Krizhevsky, Ilya Sutskever, and Ruslan
  Salakhutdinov. 2014.
\newblock Dropout: A simple way to prevent neural networks from overfitting.
\newblock \emph{JMLR}.

\bibitem[{Watanabe and Sumita(2015)}]{watanabe2015transition}
Taro Watanabe and Eiichiro Sumita. 2015.
\newblock Transition-based neural constituent parsing.
\newblock In \emph{Proc. of ACL}.

\bibitem[{Zhang et~al.(2004)Zhang, Shen, Zhou, Su, and
  Tan}]{zhang2004enhancing}
Jie Zhang, Dan Shen, Guodong Zhou, Jian Su, and Chew-Lim Tan. 2004.
\newblock Enhancing hmm-based biomedical named entity recognition by studying
  special phenomena.
\newblock \emph{Journal of Biomedical Informatics}, 37(6):411--422.

\bibitem[{Zhang and Clark(2009)}]{zhang2009transition}
Yue Zhang and Stephen Clark. 2009.
\newblock Transition-based parsing of the chinese treebank using a global
  discriminative model.
\newblock In \emph{Proc. of IWPT}.

\bibitem[{Zhou(2006)}]{zhou2006recognizing}
Guodong Zhou. 2006.
\newblock Recognizing names in biomedical texts using mutual information
  independence model and svm plus sigmoid.
\newblock \emph{International Journal of Medical Informatics}, 75(6):456--467.

\bibitem[{Zhou et~al.(2004)Zhou, Zhang, Su, Shen, and
  Tan}]{zhou2004recognizing}
Guodong Zhou, Jie Zhang, Jian Su, Dan Shen, and Chewlim Tan. 2004.
\newblock Recognizing names in biomedical texts: a machine learning approach.
\newblock \emph{Bioinformatics}, 20(7):1178--1190.

\end{thebibliography}
\bibliographystyle{acl_natbib_nourl}

\section*{Appendix}

The action constraints are listed as follows:

\begin{itemize}

\item The \textsc{Shift} action is valid only when the buffer is not empty.
\item The \textsc{Unary-X} actions are valid only when the stack is not empty.
\item The \textsc{Reduce-X} actions are valid only when the stack has two or more elements.
\item If the top element of the stack is labeled, then unary actions are not valid.
That is, \{$\textsc{X1} \rightarrow \textsc{X2}$\} is not allowed.
\item If only one of the top two elements of the stack is temporary, say \textsc{X*}, then among all reduce actions, only \textsc{Reduce-X*} and \textsc{Reduce-X} are valid.
\item If the top two elements of the stack are both temporary, then all reduce actions are not allowed.
\item If one of the elements in the stack is temporary, say \textsc{X*}, which means it is not finished, then last terminal symbol $\$$ cannot be shifted until it is reduced to \textsc{X}.

\end{itemize}

\end{document}